\documentclass[10pt,twocolumn,letterpaper]{article}

\usepackage{iccv}
\usepackage{times}
\usepackage{epsfig}
\usepackage{graphicx}
\usepackage{amsmath}
\usepackage{amssymb}
\usepackage{booktabs}
\usepackage{arydshln}
\usepackage{enumitem}
\usepackage[dash,dot]{dashundergaps}

\usepackage[pagebackref=true,breaklinks=true,letterpaper=true,colorlinks,bookmarks=false]{hyperref}

\usepackage[capitalize]{cleveref}
\crefname{section}{Sec.}{Secs.}
\Crefname{section}{Section}{Sections}
\Crefname{table}{Table}{Tables}
\crefname{table}{Tab.}{Tabs.}
\iccvfinalcopy 

\def\iccvPaperID{9326} 

\ificcvfinal\fi

\def\authornote#1#2#3{{\textcolor{#2}{{\small[#1: #3]}}}}
\iftrue
\newcommand{\cfangyin}[1]{\authornote{Fangyin}{blue}{#1}}
\else
\newcommand{\cfangyin}[1]{}
\fi
\begin{document}

\title{Clutter Detection and Removal in 3D Scenes with View-Consistent Inpainting}

\author{Fangyin Wei
\and
Thomas Funkhouser\\
{\qquad\qquad Princeton University}
\and
Szymon Rusinkiewicz
}

\maketitle
\ificcvfinal\thispagestyle{empty}\fi

\begin{abstract}
Removing clutter from scenes is essential in many applications, ranging from privacy-concerned content filtering to data augmentation. 
In this work, we present an automatic system that removes clutter from 3D scenes and inpaints with coherent geometry and texture. We propose techniques for its two key components: 3D segmentation based on shared properties and 3D inpainting, both of which are important problems.
%
%
The definition of 3D scene clutter (frequently-moving objects) is not well captured by commonly-studied object categories in computer vision. 
To tackle the lack of well-defined clutter annotations, we group noisy fine-grained labels, leverage virtual rendering, and impose an instance-level area-sensitive loss.
Once clutter is removed, we inpaint geometry and texture in the resulting holes by merging inpainted RGB-D images. 
This requires novel voting and pruning strategies that guarantee multi-view consistency across individually inpainted images for mesh reconstruction. 
Experiments on ScanNet and Matterport3D dataset show that our method outperforms baselines for clutter segmentation and 3D inpainting, both visually and quantitatively. 
Project page: ~\href{https://weify627.github.io/clutter/}{https://weify627.github.io/clutter/}.

\end{abstract}


\section{Introduction}
\label{sec:intro}
With the proliferation of RGB-D cameras, we can imagine a day when people can quickly scan their rooms with a phone and upload the reconstructed 3D model to a website for subleasing. It is more desirable to show an attractive, clean room without pillows and kitchenware scattered all around. However, people prefer not to spend hours cleaning them up before scanning. This is when a tool for automatic scene clutter removal would come in handy. Such a scene editing tool can be applied to a wide range of tasks from automatic content and privacy filtering to data augmentation (by adding new objects to cleaned scenes).



In this paper, we investigate how to reconstruct a scene without clutter, starting from a set of indoor RGB-D scans. This task consists of two steps: clutter segmentation and 3D inpainting, both of which present new challenges.

\begin{figure}
    \centering
 
    \includegraphics[trim={0.0cm 0 0 0}, clip,width=1\linewidth]{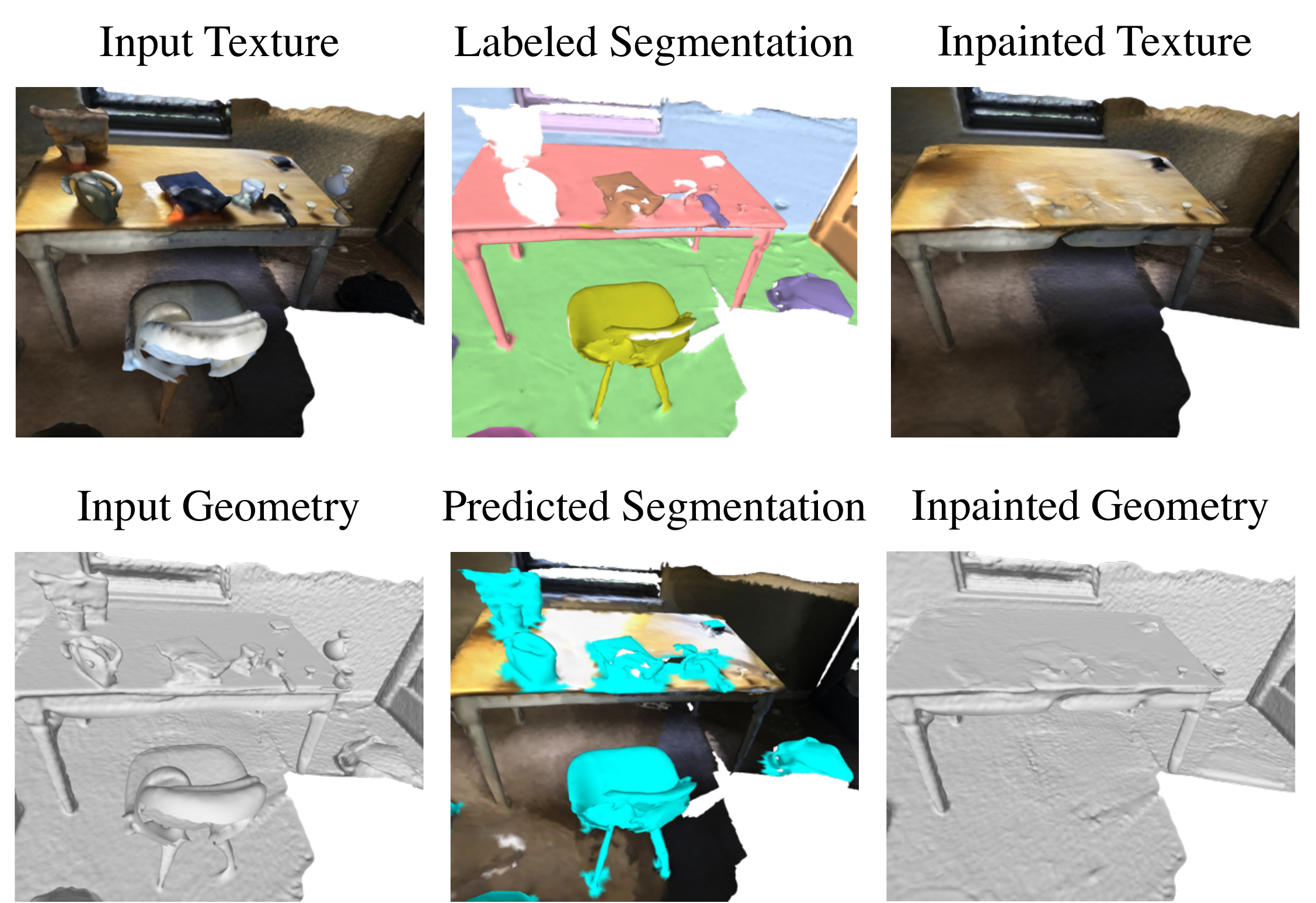}
    \caption{\textbf{Clutter detection and removal in 3D Scenes with view-consistent inpainting.}
    Beginning from an input scene (left), our first task is to predict clutter regions.  Traditional semantic segmentation datasets and methods \cite{dai2017scannet} (top center) were not designed for this problem, and contain missing or incorrect labels (especially for small objects) as well as semantic categories that only sometimes correspond to clutter.  In contrast, our method (bottom center) correctly identifies clutter segments, which we then remove and inpaint with coherent texture and geometry (right).
    }
    \label{fig:teaser}
\end{figure}
Thanks to large-scale 3D indoor scene datasets~\cite{chang2017matterport3d,dai2017scannet,armeni20163d,rozenberszki2022language}, data-driven methods have led to remarkable success in 3D scene segmentation~\cite{hu-2021-bidirectional,nekrasov2021mix3d,kundu2020virtual}. However, existing annotations are insufficient for clutter segmentation for two reasons. First, ambiguity exists even in the most fine-grained datasets to date (around 550 raw categories for datasets~\cite{chang2017matterport3d,dai2017scannet}, 200 for benchmark~\cite{rozenberszki2022language}). For example, some objects from the category \textit{whiteboard} are portable and belong to clutter, while others are installed on the wall and not clutter. Such intra-category ambiguity is widespread: consider \textit{light}, \textit{seat}, \textit{speaker}, \textit{decoration}, \etc, all of which exist in both fixed (non-clutter) and mobile (clutter) variations. Secondly, the quality of annotations is often suboptimal, with some clutter objects being unannotated or mislabeled as non-clutter, as shown in Fig.~\ref{fig:teaser}, top center. Considering the mesh resolution and the time required to label a large number of small objects, it is almost impossible to annotate clutter precisely and exhaustively in a large dataset.


The task of 3D inpainting aims to synthesize a visually and geometrically plausible completion for the missing regions of a 3D scene. 
While some prior work~\cite{setty2018example,wang20203d,mittal2021self} inpaints smaller holes from incomplete scans or occlusions, we focus on larger holes created by object removal, which exacerbates the difficulty of inpainting due to complex geometry and texture. A naive solution is to run 3D hole-filling or mesh reconstruction algorithms on the remaining points. However, these algorithms only take into account local geometry and low-level texture information, whereas in many cases, semantic understanding is required for plausible completion. For example, when some clutter covering a table corner is removed, the inpainted scene is expected to recover the corner rather than just filling the hole and leaving the corner missing. Directly working on the 3D scene entails dealing with entangled semantics and complex geometry at the same time, which may result in low resolution and many artifacts in the output~\cite{dai2021spsg,jheng2022free}. We propose to instead inpaint the RGB-D sequences and then reconstruct the final mesh. This disentangles the problem into semantically plausible inpainting and geometry filling. The former can be solved more easily with existing advanced 2D inpainting methods and potentially more powerful than methods trained only on limited 2D data.
 

In this work, we endeavor to address the above challenges and propose an automatic system for scene clutter removal and 3D inpainting. 
 To more effectively segment small clutter objects, we design  an area-sensitive loss to force attention to small clutter objects which typically receive limited loss signals. We adopt a deep architecture that takes in both RGB images and 3D points as input  and render virtual views as the 2D input to gain large surface coverage.
Different from prior work that directly predicts inpainting in 3D after identifying objects in 3D, we project the object masks onto RGB-D images and perform image inpainting and image-guided depth completion. To enforce cross-frame consistency for a better quality of 3D reconstruction, we run novel consistency voting and pruning across frames. We loop back to image inpainting until all missing regions have been filled. The resulting collection of consistently inpainted RGB-D images can be merged into a final, clean scene.

We conduct extensive ablation studies to validate the system designs on the ScanNet~\cite{dai2017scannet} and Matterport3D~\cite{chang2017matterport3d} datasets. We also compare with 3D segmentation and inpainting baselines and show the effectiveness of the proposed model.
In summary, our contributions are:
\begin{itemize}[itemsep=0.1pt,topsep=1pt,leftmargin=*]
    \item An automatic system that solves the novel task of scene clutter removal and inpainting.
    \item A 3D segmentation method with area-sensitive loss that can better segment small objects even with noisy data.
    \item A 3D inpainting method with iterative view-consistent RGB-D inpainting and outperforms prior methods. 
\end{itemize}







\section{Related Work}
\label{sec:relatedwork}
\paragraph{3D Semantic Segmentation}
Many prior works have studied how to segment 3D point clouds into semantic classes.  This work focuses on identifying
specific types of object categories~\cite{dai2017scannet,chang2017matterport3d,armeni20163d}, and thus generally works well when label sets are small \cite{hu-2021-bidirectional,nekrasov2021mix3d,kundu2020virtual}, but not so well when there are many possible classes and few examples are available for most in the training set~\cite{rozenberszki2022language}. In this paper, we address the more abstract tasks of identifying clutter \vs not, an attribute that does not directly align with object category labels. 


\paragraph{RGB Image Inpainting}
Traditional single-view methods~\cite{efros1999texture,efros2001image,ballester2001filling,barnes2009patchmatch} such as patch-based or diffusion-based work often achieve realistic results but lack high-level image understanding. 
Since the seminal work~\cite{pathak2016context} that introduced adversarial training and encoder-decoder structure into this task, various techniques have been proposed. These include multi-stage generation~\cite{nazeri2019edgeconnect,ren2019structureflow}, gated convolution~\cite{yu2019free}, recurrent methods~\cite{li2020recurrent,zhang2018semantic}, \etc.
More recent work begins to focus on more challenging settings such as pluralistic inpainting~\cite{zheng2019pluralistic,wan2021high} and large hole filling~\cite{zhao2021large,li2022mat,suvorov2022lama}. Specifically, LaMa~\cite{suvorov2022lama} uses fast Fourier Convolutions and extremely large masks for training. It achieves high-fidelity results with relatively low training and inference cost. Single image 3D photography aims to synthesize novel views using image-based rendering techniques~\cite{jampani2021slide,shih20203d,kopf2020one}. In some sense, it can be regarded as filling in small holes in other views from a single image.
 



\paragraph{Depth, Multi-View, and Mesh Inpainting}
Recent deep learning-based depth-completion methods have made great success on both supervised~\cite{park2020non,hu2021penet,cheng2020cspn++,yang2019dense} and unsupervised~\cite{ma2019self,wong2021unsupervised,lopez2020project} learning. In particular, NLSPN~\cite{park2020non}
performs non-local depth propagation using deformable convolutions~\cite{zhu2019deformable}. 
Multi-view inpainting is commonly used as a technique for image-based rendering~\cite{thonat2016multi,li2018multi,yan2022image,philip2018plane,mori20183d}. These methods and our method both involve 3D-aware image inpainting. However, we aim to obtain inpainted mesh, whereas multi-view inpainting aims to render new views,~\ie, the cross-frame consistency can be optimized as one goal but not guaranteed.
The word of ``inpainting'' has been overloaded for shape completion~\cite{horry1997tour,setty2018example,wang20203d,mittal2021self}.  For instance, the example-based method~\cite{setty2018example} aims to fill in small holes from incomplete scans. These methods only focus on geometry. Surface reconstruction algorithms~\cite{Peng2021SAP, kazhdan2006poisson} can only create surface between close enough points, therefore cannot generate new surfaces to fill big holes.


\paragraph{3D Scene Inpainting}
3D scene inpainting aims to fill in plausible geometry (and texture) for missing parts of a 3D scene. Earlier work only completes geometries~\cite{dai2018scancomplete,dai2020sgnn}, and recently 3D inpainting of both geometry and texture has gained more interests~\cite{dai2021spsg,jheng2022free,mirzaei2023spin,weder2023removing}. TSDF-based methods~\cite{dai2021spsg,jheng2022free} use one forward pass to predict the scene completion. They are good at recovering dominant structures (\eg flat walls and floors) but often struggle with more challenging structures. Our mesh-based method instead iteratively refines the scene by disentangling geometry and texture. Concurrent to our method, two NeRF-based methods~\cite{mirzaei2023spin,weder2023removing} also employ iterative refinement for scene object removal. One key difference is how multi-view inconsistency is addressed. We design pruning and voting techniques to select consistent views; Weder~\etal~\cite{weder2023removing} learn a per-image confidence to re-weight the NeRF losses; SPIn-NeRF~\cite{mirzaei2023spin} does not select view and does not enforce consistency. Another difference is how depth is inpainted. While~\cite{mirzaei2023spin,weder2023removing} inpaint depths only from masked depths, we perform RGB-guided depth inpainting and better align the inpainted depths with inpainted RGB images.





\section{Method}
\label{sec:method}
\begin{figure*}
    \centering
    \includegraphics[width=1\linewidth]{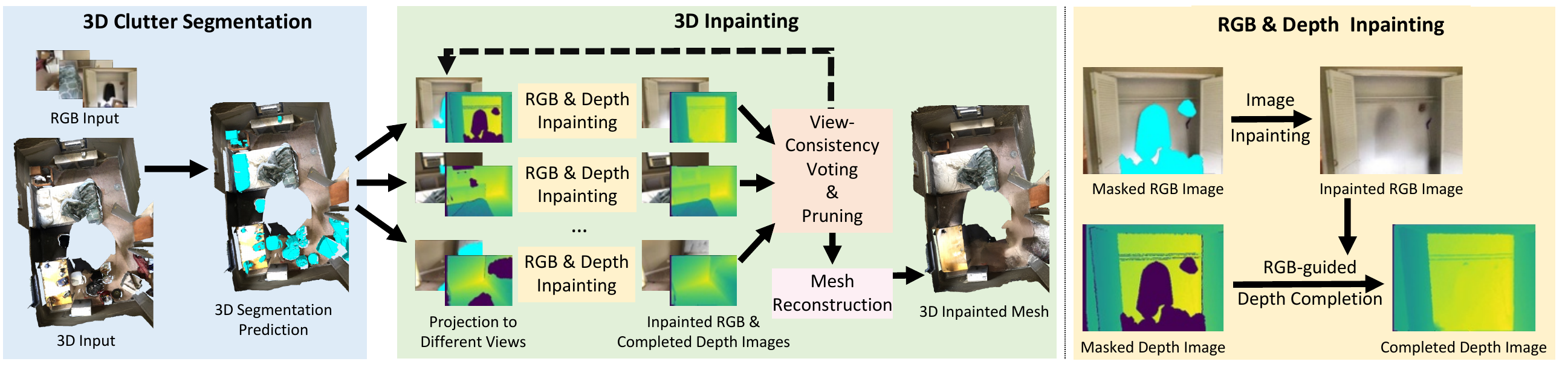}
    \caption{\textbf{Method overview.} Given an RGB-D reconstruction and the RGB frames, we first predict 3D clutter segmentation. The prediction is then projected in different camera views to mask RGB and depth images. Next, we inpaint RGB image which we use to guide depth completion. Afterwards, we run consistency pruning and voting for all RGB-D frames. Any inpainting with remaining missing regions is sent back to RGB and depth inpainting again. The loop continues until all missing regions are filled. As last, we obtain the final inpainted scene mesh through mesh reconstruction.}
    \label{fig:overview}
\end{figure*}
Given inputs of an RGB-D sequence and its reconstructed mesh of an indoor scene, our method outputs a 3D scene without clutter (as shown in Fig.~\ref{fig:overview}). The first step is to predict the 3D clutter segmentation (Sec.~\ref{subsec:clutter-seg}). We then project each clutter segment into each camera view to mask out clutter regions in the color and depth images. This is followed by color image inpainting, the result of which guides depth completion (Sec.~\ref{subsec:single-inpaint}). Once all RGB-D frames are inpainted, we run consistency pruning and voting. If there remain missing regions from clutter removal, the consistency-checked inpaintings are sent back to image inpainting and depth completion to fill the remaining holes (Sec.~\ref{subsec:multi-inpaint}). The loop continues until all missing regions are filled. As a final step, we run Poisson surface reconstruction~\cite{kazhdan2006poisson} on the consistency-guaranteed inpaintings to obtain the final inpainted scene mesh (Sec.~\ref{subsec:mesh-inpaint}).
\subsection{Clutter Segmentation}
\label{subsec:clutter-seg}

Our focus is to propose a new type of scene segmentation based on shared properties (\eg across clutter) rather than common benchmark semantic categories. Therefore, a concrete clutter definition is crucial to test our method  although the precise form of the definition can be flexible (discussed in supplementary).
Once clutter is defined, we group all raw categories (549 for ScanNet~\cite{dai2017scannet}) into clutter/non-clutter/mixed. To reduce ambiguity, we exclude mixed categories and only use other categories as coarse supervision.
Note that due to the heterogeneity of clutter, 
most clutter categories have fewer than 20 instances each and the total instance count of clutter is smaller than that of non-clutter.
\subsubsection{Area-Sensitive Loss}
3D segmentation models typically exploit Cross Entropy loss to supervise 3D predictions.
However, directly applying the standard Cross Entropy loss does not train a model for our task with a performance similar to~\cite{hu-2021-bidirectional}. A unique feature of clutter compared to general objects is that a larger portion of clutter objects are small in size. However, a standard per-vertex loss focuses too much on larger instances. To draw the model's attention back to small objects, we modify the standard Cross Entropy loss into the form:
\[
L_{CE(3d)}^* = \frac{1}{N_v}\sum_{c\in C}\sum_{i\in O_c}\Bigl(w_i\sum_{j\in V_i} y_{(c,j)} log(p'_{(c,j)})\Bigr),
\]
where $N_v$ is the total number of vertices in the scene, $C$ is the set of all classes (clutter and non-clutter), $O_c$ is the set of instances of class $c$ in the current scene, $V_i$ is the set of vertices of the object $i$, $y_{(c,j)}$ and $p'_{(c,j)}$ are ground truth labels and predicted class probabilities for vertex $j$, respectively. In particular, $w_i$ is our introduced weight to balance instance size and is formulated as:
\[
w_i = \Bigl(\frac{median(N)}{n_i}\Bigr)^k, n_i\in N,
\]
where $N$ is a list of the vertex counts for all objects in the scene, $n_i$ is the vertex count for object $i$, and $k\ge0$ is a modulating factor to control how strongly the weight is applied. A smaller $n_i$ (fewer vertices in object instance $i$) increases the weight $w_i$ and loss for object $i$. A larger $k$ increases $w_i$ and loss penalty for smaller object instances. 
In particular, when $k=0$ and $w_i=1$, $L_{CE(3d)}^*$ takes the same form as Cross Entropy loss.

It is worth pointing out that our area-sensitive loss may look superficially similar to median frequency loss~\cite{eigen2015predicting} or balanced Cross Entropy~\cite{jadon2020survey,cui2019class}. However, we are fundamentally different. While other losses are aimed to address the class imbalance, our loss is on an instance level to better segment small objects -- object instances with small surface area are weighted more regardless of the total surface area for the class. In particular, not all clutter instances are small (there are some easy and big-sized categories), and not all non-clutter instances are big. We expect our loss to be helpful for small objects across classes. A class-balanced loss in our binary segmentation task would simply be weighting one class uniformly larger. We found empirically that this does not improve the results of our task.



\subsubsection{Implementation and Virtual View Rendering}
Our focus is to design effective losses for heterogeneous small objects. We opt for BPNet~\cite{hu-2021-bidirectional} as the backbone 3D segmentation model. It contains 2D and 3D sub-networks connected by bidirectional feature projection, which takes as input color images and point clouds, respectively. The loss function 
$L_{seg} = L_{CE(3d)}^* + \lambda L_{CE(2d)}^*$
is a weighted sum of area-sensitive losses on 3D and 2D predictions, balanced by a weight $\lambda$.
We follow ~\cite{hu-2021-bidirectional} and input 3 views at a time. However, three originally captured images only cover a small surface of the entire scene, and increasing the number of input views drastically consumes the memory. Inspired by~\cite{kundu2020virtual}, we replace the originally captured images with virtual renderings for which we can set camera parameters to increase observed context.


\subsection{3D Inpainting}
3D inpainting is a recently emerging task~\cite{dai2021spsg,jheng2022free}, and one key challenge in 3D inpainting is the intricate interplay between geometry and semantics (or appearance). To address this issue, our key idea is to leverage intermediate 2D representations that disentangles textured 3D into RGB images (appearance) and geometry, which can be more easily inpainted and then used for multi-view 3D reconstruction. Notably, this novel 2D approach to the 3D inpainting problem deviates from prior work that relies on 3D convolutions for direct inpainting in 3D \cite{dai2021spsg,jheng2022free}. In the following, we will first briefly overview the proposed pipeline before introducing new techniques to tackle the challenges.

\subsubsection{Single-Frame Inpainting}
\label{subsec:single-inpaint}
We first project the clutter predictions from a 3D mesh onto each view to create a 2D mask for RGB and depth images. One benefit of our proposed 2D approach is to leverage the powerful 2D synthesis models. This also potentially improves generalization ability especially when faced with limited 3D data (compared to image data).

\paragraph{Color Image Inpainting}

We use LaMa~\cite{suvorov2022lama} with released pre-trained weights to inpaint each color image. The model trained on diverse data generalizes well to indoor captured images. However, in cases where multiple completions are plausible (especially for big masks), the model tends to fill in different contents for the same region from different views. This also leads to inconsistent depth completion and will be addressed in Sec.~\ref{subsec:multi-inpaint}. 
\paragraph{Depth Completion}
Existing depth completion methods~\cite{park2020non,hu2021penet,cheng2020cspn++,yang2019dense,ma2019self,wong2021unsupervised,lopez2020project} try to predict complete depth map from uniformly and sparsely sampled pixels,~\eg, by using local propagation. However, we need to complete dense maps with (potentially big) holes without local information. We adopt NLSPN~\cite{park2020non} that performs non-local depth propagation for image-guided depth completion. Since there's no annotated data for our task, we design a method that creates realistic holes on originally clean depth maps to generate supervision:

We assume $m_1$ is a coarse clutter mask of a captured depth map $d_1$. We use the clutter mask $m_2$ from a second view to mask out regions on $d_1$. In the meantime, we make sure that the regions originally masked as clutter in $m_1$ are not masked out. By copying and pasting $m_2$ from another view, we make sure that the masks during training have realistic boundaries of clutter objects. By not masking clutter from $m_1$, we guarantee that the depth completion model does not need to hallucinate clutter objects out of the hole areas.


\subsubsection{View-Consistency Inspired Refinement}
\label{subsec:multi-inpaint}
The tremendous advances in image inpainting and depth completion have enabled high-quality inpainting of individual frames. However, due to the pluralistic nature of image inpainting and the accuracy limits of each model, inconsistency occurs across frames (Fig.~\ref{fig:ablation-inp}). Such inconsistency will cause many noises in the mesh reconstruction.  In this section, we discuss the unique characteristics of the task and explain how we enforce consistency based on the observations. 

\paragraph{Single-frame Consistency Pruning}
Inherent in the task of object removal, the inpainted depth $ d_{pre}$ should always be greater than the original captured depth $d_{cap}$ at the same region. Here, the region includes each pixel $p$ and each connected mask region $i$.
Based on this observation, we have
\[
d_{con}^1(p) = 
\begin{cases}
  d_{pre}(p) & \text{if}\ \ \ d_{cap}(p) < d_{pre}(p),\\
  0  & \text{else;}
\end{cases}
\]
\[
d_{con}^2(i) = 
\begin{cases}
  d_{con}^1(i) & \text{if}\ \ \ d_{cap}(i) < d_{pre}(i),\\
  0  & \text{else,}
\end{cases}
\]
where $d_{con}^1$ and $d_{con}^2$ are the depth outputs from consistency checks for pixels and connected regions, respectively. Since our focus is to fill holes from object removal rather than the cracks in captured depth images caused by moving sensors, we mask out any pixel in $d_{pre}$ that is empty in $d_{cap}$. We also drop inpainted regions whose area is over half the entire image, because inpaintings from limited input pixels are highly unreliable.
\paragraph{Cross-frame Consistency Pruning}
In the same spirit as space carving, the inpainted depth $d^{(s)}_{pre}$ from source view $s$, when warped to a target view $t$, cannot occlude captured depth in that view $d_{cap}^{(t)}$. This means at each pixel $p$:
\[
d_{con}^{3(s)}(p) = 
\begin{cases}
  d_{con}^{2(s)}(p) & \text{if}\ \ \ d_{cap}^{(t)}(p) < \text{Warp}^{s-t}\Bigl(d^{(s)}_{pre}(p)\Bigr), \\
  0  & \text{else,}
\end{cases}
\]
where $d_{con}^{3(s)}$ is the depth output under view $s$ from cross-frame consistency pruning. Warp$^{s-t}(\cdot)$ is the transformation from view $s$ to view $t$. The transformation can be calculated given camera intrinsics and poses for two views. This is performed for each view pair $(s, t)$ with $s,t\in\{0, \dots, N_s-1\}$, where $N_s$ is the sequence length. 

\paragraph{Cross-frame Consistency Voting}
So far we have pruned some regions base on consistency to captured depths. However, there remains inconsistency among inpainted regions themselves. The causes include: at the same region from different views, (1) color images are inpainted with different contents or (2) completed depths are different even color image inpaintings are the same. To address this, we warp inpainted regions of all other views to view $s$ using the predicted depth. Then we consider all warped depth values that fall into pixel $p$. The depth value at pixel $p$ for view $s$ is only valid if the count of projected depths within a distance threshold $\alpha$ from $d^{(s)}_{pre}(p)$ is more than $\beta\%$ of the total count of projected depths falling in at $p$. Thus we have
\[
d_{con}^{4(s)}(p) = 
\begin{cases}
  d_{con}^{3(s)}(p) & \text{if}\ \ \ r_\alpha >\beta\%, \\
  0  & \text{else,}
\end{cases}
\]
\[
r_\alpha=\frac{\sum_{t}\mathbb{I}\Bigl(| d_{pre}^{(s)}(p) - \text{Warp}^{t-s}\Bigl(d^{(t)}_{pre}(p)\Bigr)|<\alpha\Bigr)}{\sum_{t}\mathbb{I}\Bigl( \text{Warp}^{t-s}\Bigl(d^{(t)}_{pre}(p)\Bigr)>0\Bigr)}, 
\]
where $t\in\{0, \dots, N_s-1\} \setminus \{ s\}$, $d_{con}^{4(s)}$ is the depth output at view $s$ from cross-frame consistency voting, and $\mathbb{I}(\cdot)$ is a binary indicator function with $\mathbb{I}(x)=1$ if $x$ is true else $0$.
\paragraph{Iterative Refinement and Efficiency}
These consistency checks create new holes in $d_{con}^{4(s)}$. We mask inpainted color images and depth maps at the current step with these new holes and start over 3D inpainting with the new masked RGB-D images. The single-frame inpainting and consistency refinement are iteratively performed until all holes are filled. The complexity of the cross-frame pruning or voting is $O(N_s^2)$. We implement a batched warping function in PyTorch with GPU support, which takes about 7 minutes for a 200-frame sequence on an NVIDIA TITAN GPU.
\subsubsection{Mesh Reconstruction}
\label{subsec:mesh-inpaint}
To obtain the final inpainted 3D scene, we run Poisson surface reconstruction~\cite{kazhdan2006poisson} on the inpaintings from Sec.~\ref{subsec:multi-inpaint}. We use a maximum of 16 million points and depth values smaller than 3.5 meters from the inpaintings. The maximum depth of the tree for surface reconstruction is set to 10.

\section{Experiments}
\label{sec:experiment}

\subsection{Experiment Setup}

\paragraph{Dataset} 
We use ScanNet dataset~\cite{dai2017scannet} with official split. Since clutter labels from grouped raw categories are very noisy, the original test set does not provide accurate evaluation.  We manually label 8 representative test scenes with diverse settings using the annotation tool from~\cite{dai2017scannet}. Evaluations on this clean labeled set truly reflect performance. Please refer to supplementary materials for more details. 
\paragraph{Implementation Details} 
 We set segmentation voxel size to 5 cm, with loss balancing weight $\lambda$ and modulating factor $k$ empirically set to 0.3 and 1 respectively.

 We set $\alpha=0.05, \beta\%=30\%$ for the cross-frame consistency voting. 
\subsection{Baselines and Evaluation Metric}
\paragraph{Clutter Segmentation}
We compare with a baseline that trains the original multi-class BPNet~\cite{hu-2021-bidirectional} for segmentation of 549 raw categories and then merges predictions based on our clutter/non-clutter grouping. All the other settings are the same as our model. We evaluate 3D mean IoU of both methods on two test sets using virtual views.
\paragraph{3D Inpainting}
We compare with the hole-filling algorithm ``Close Holes'' from MeshLab~\cite{meshlab,LocalChapterEvents:ItalChap:ItalianChapConf2008:129-136} and Poisson surface reconstruction (PSR)~\cite{kazhdan2006poisson} with the same setting as in Sec.~\ref{subsec:mesh-inpaint}. We further compare with two recent textured scene completion methods (SPSG~\cite{dai2021spsg} and FF~\cite{jheng2022free}). Since they were both developed on Matterport3D dataset~\cite{chang2017matterport3d}. We also retrain our method on Matterport3D and compare with their official code and model weights. 


\subsection{Results}
\subsubsection{Full Pipeline}
\begin{figure*}
    \centering
    \includegraphics[width=1\linewidth]{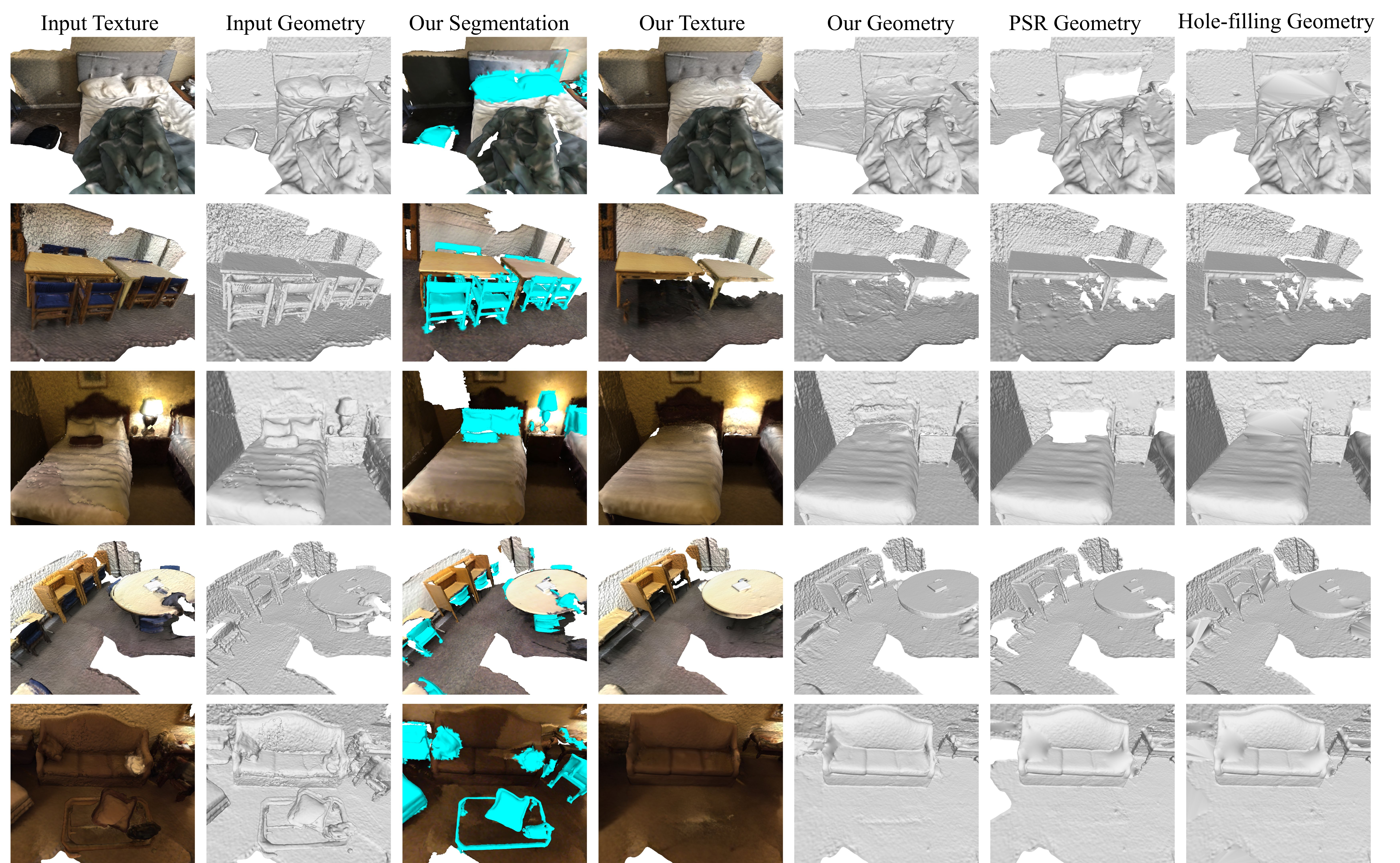}
    \caption{\textbf{Results for automatic clutter removal and 3D inpainting.}  Each row shows a rendered view of an input 3D scene, together with our clutter segmentation and a rendered view of the scene with inpainted color and geometry.  We compare the latter to removing clutter and filling the resulting holes with either Poisson Surface Reconstruction (PSR) or a triangulation-based hole-filling algorithm.}
    \label{fig:main-result}
\end{figure*}
\begin{table}[t]
\caption{\textbf{Comparison of clutter segmentation methods.} We evaluate IoU and compare our method with the BPNet baseline.}
\label{tab:3dseg-comparison}
\addtolength{\tabcolsep}{-1pt}
\centering
\resizebox{1\linewidth}{!}{
\begin{tabular}{lcccccc}
\specialrule{.15em}{.1em}{.1em}
Method & \multicolumn{3}{c}{Original Test Set (noisy)} &\multicolumn{3}{c}{Manual Test Set (clean)} \\
      & IoU(NC) & IoU(C) & \multicolumn{1}{c}{mIoU} & IoU(NC) & IoU(C) & mIoU \\
\hline
BPNet~\cite{hu-2021-bidirectional}&0.87&0.43&0.65&\textbf{0.85}&0.35&0.60\\
Ours &\textbf{0.90}& \textbf{0.59}& \textbf{0.75}&0.84&\textbf{0.58}&\textbf{0.71} \\
\specialrule{.1em}{.05em}{.05em}
\end{tabular}
}
\end{table}

\begin{figure}
    \centering
    \includegraphics[trim={0.5cm 0 0 0}, clip, width=1\linewidth]{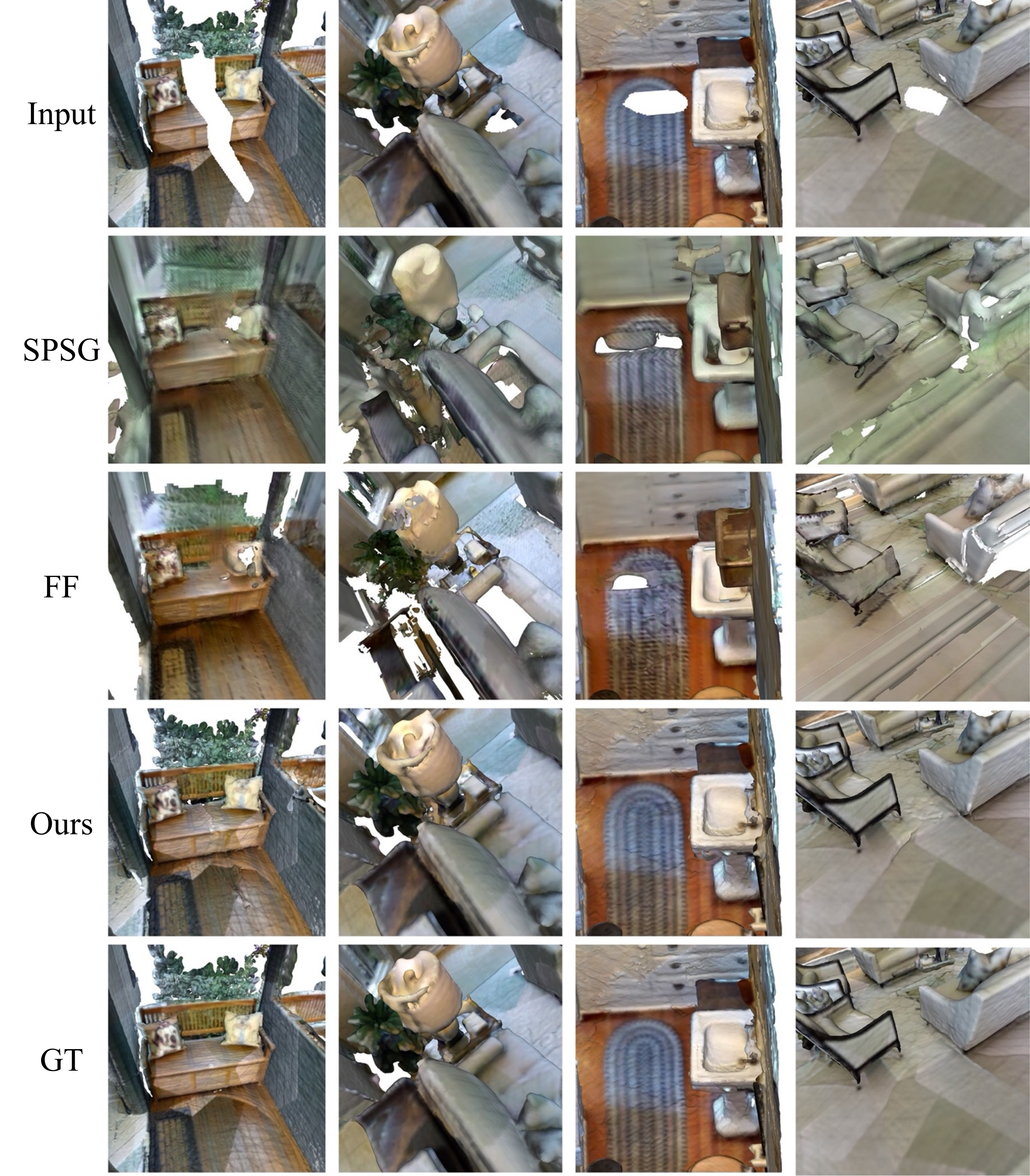}
    \caption{\textbf{3D inpainting comparison.} We compare our method with two 3D scene inpainting (completion) baselines (SPSG~\cite{dai2021spsg} and FF~\cite{jheng2022free}). The first row shows input incomplete scenes (holes created by dropping clutter objects onto the scene), followed by three methods' outputs and the GT scene without synthetic holes at the bottom. While prior work suffers from low resolution, our approach can keep the same resolution as input and is better at completing larger holes and rarer scenes.}
    \label{fig:matterport-inp}
\end{figure}

In Fig.~\ref{fig:main-result}, we show qualitative results for our entire pipeline of object removal and 3D inpainting, where we compare with baselines of 3D inpainting. The middle three columns show our predicted 3D clutter segmentation, inpainted texture and geometry. On the right, we compare with the geometry from Poisson surface reconstruction (PSR)~\cite{kazhdan2006poisson} and a hole-filling algorithm. 

The third column shows that our segmentation method not only accurately predicts bigger and regularly-shaped categories such as coffee tables (last row) and chairs, but also performs well on objects with irregular shapes and appear at very low frequency. We also observe that the segmentation baseline often miss small objects. Detecting these small objects is nontrivial due to the large annotation noise in the training set (Fig.~\ref{fig:teaser}). Our method's successful detection mostly benefits from our area-sensitive loss, which we will thoroughly study in Sec.~\ref{subsec:3dseg-abl}.

The fourth column illustrates that our method can fill in the missing regions with consistent texture. In the fourth row, for example, dark colors from the clutter on the desk all disappear after inpainting. We attribute this to the powerful semantic understanding from image inpainting. However, there still exist inpainted texture artifacts, and a considerable number of them are caused by illumination. Taking the third row as an example, we can successfully segment the entire lamp but the light it casts stays, which is not ideal. 

The fifth column shows that our method is better at closing holes caused by object removal and completing with geometrically plausible surfaces. For instance, in the fourth row, the desk on the left has a big missing region, caused by the chair's occlusion. PSR fails to fill this large hole and the hole-filling algorithm fills the hole incorrectly by connecting the desktop with desk legs. In contrast, our method not only fills the entire hole but also recovers the missing part of the desktop. Similar phenomena can also be found in other scenes, \eg pillow regions on the bed in the first row and on the sofa in the last row.

\subsubsection{Segmentation}
We report in Tab.~\ref{tab:3dseg-comparison} the IoU evaluations of the baseline and our method. It shows that our method outperforms the compared BPNet~\cite{hu-2021-bidirectional} with multi-category segmentation. This is not too surprising, since the compared model was originally applied for tasks with 20 categories~\cite{hu-2021-bidirectional} while there are 549 categories in our experiment. As noted by prior work~\cite{rozenberszki2022language}, the large number of class categories results in severe class imbalance. In summary, the comparison illustrates that the problem of clutter segmentation cannot be solved with a straightforward fine-grained segmentation model. More analysis on the importance of each component introduced to segmentation can be found in Sec.~\ref{subsec:3dseg-abl}.

\subsubsection{3D inpainting}
\paragraph{Qualitative Evaluation}
We compare with the most recent textured 3D scene inpainting methods in Fig.~\ref{fig:matterport-inp}. Both methods use 3D convolutions to directly predict TSDF from incomplete input volume. They suffer from low resolution which is a typical weakness of methods using volumetric representation. In contrast, our method outputs high-resolution mesh. In the second column, SPSG~\cite{dai2021spsg} and FF~\cite{jheng2022free} fail to complete the hole on the couch, whereas our method recovers color and geometry consistent with the input. Another advantage of our proposed 2D approach is robustness and better generalization. For example, the first column in Fig.~\ref{fig:matterport-inp} shows an outdoor scene example with large holes, which is rarely seen during training. However, since our 2D inpainting modules have seen more diverse training example, our method can also work well on uncommon examples.

\begin{table}[t]
\caption{\textbf{Comparison of 3D inpainting methods.} We quantitatively compare with baselines using both 2D and 3D metrics.}
\label{tab:3dinp-comparison}
\centering
\resizebox{1\linewidth}{!}{
\begin{tabular}{lcccccc}
\specialrule{.2em}{.1em}{.1em}
Method & L1($\downarrow$)& L2($\downarrow$)& PSNR($\uparrow$)& SSIM($\uparrow$)&LPIPS($\downarrow$)&CD(cm)($\downarrow$)\\
\hline
SPSG~\cite{LocalChapterEvents:ItalChap:ItalianChapConf2008:129-136}& 0.077& 0.139&16.810&0.696& 0.419 & 21.354 \\
FF~\cite{kazhdan2006poisson}&0.038& 0.071& 21.762& 0.861& 0.213&1.781\\
Ours &\textbf{0.018}& \textbf{0.033}&\textbf{23.673}& \textbf{0.918}& \textbf{0.151}&\textbf{0.621}\\
\specialrule{.1em}{.05em}{.05em}
\end{tabular}
}
\end{table}

\paragraph{Quantitative Evaluation}
To quantitatively compare our method with baselines, we create a dataset with synthetically created holes (details in the supplementary). To evaluate the inpainted mesh texture, we report L1, L2, PSNR, SSIM, and LPIPS on the renderings of inpainted meshes, following common practice in image inpainting~\cite{yu2019free,zheng2019pluralistic}. To evaluate inpainted mesh geometry, we compare the Chamfer Distance between the original surface and the reconstructed surface. We report the evaluation results in Tab.~\ref{tab:3dinp-comparison}.  We can see our proposed method outperforms baselines on all metrics. This indicates that our method produces better texture and geometry, which has been shown in Fig.~\ref{fig:main-result} and Fig.~\ref{fig:matterport-inp}.

\subsection{Ablation Studies}

\subsubsection{Clutter Segmentation}
\label{subsec:3dseg-abl}
\begin{table}[t!]
\caption{\textbf{Ablations on clutter segmentation.} We study the effect of our area-sensitive loss with varying $k$. We also compare the influence of using rendered($r.$)/captured($c.$) 2D views and new($n.$)/original($o.$) camera parameters for training. (a-e) are evaluated on $r./n.$ views, and (f-h) are evaluated on $c./o.$ views. Best/second-best/third-best results among (a-e) are \textbf{bold}, \underline{solid underlined}, and \dashuline{dashed underlined}, respectively. }
\label{tab:3dseg-abl}
\addtolength{\tabcolsep}{-4pt}
\centering
\resizebox{1\linewidth}{!}{
\begin{tabular}{lccccccccccc}
\specialrule{.15em}{.1em}{.1em}
ID&$k$ &2D Type& \multicolumn{4}{c}{Original Test Set (noisy)} &\multicolumn{4}{c}{Manual Test Set (clean)} \\
&&/Camera    & IoU(NC) & IoU(C) &Pre.& \multicolumn{1}{c}{Rec.} & IoU(NC) & IoU(C) & Pre.&Rec. \\
\hline
(a) &0 &$r./n.$&\textbf{0.94}&\textbf{0.67}&\textbf{0.85}&\dashuline{0.76}&\dashuline{0.83}&0.48&\textbf{0.91}&0.50\\
(b)&0.5&$r./n.$&\underline{0.93}&\underline{0.65}&\underline{0.79}&\underline{0.79}&\textbf{0.85}&\dashuline{0.55}&\underline{0.88}&0.59
\\
(c)&1.0&$r./n.$&\dashuline{0.90}&\dashuline{0.59}&\dashuline{0.67}&\textbf{0.83}&\underline{0.84}&\underline{0.58}&\dashuline{0.82}&\dashuline{0.66}
\\
(d)&1.5&$r./n.$ & 0.86&0.51&0.57&\textbf{0.83}&\dashuline{0.83}&\textbf{0.59}&0.76&\textbf{0.73}
\\
(e)&2.0&$r./n.$&0.79&0.39&0.45&0.75&0.78&0.51&0.65&\underline{0.70}\\
\hdashline
(f)&1.0&$c./o.$&0.92&0.62&0.76&0.77&0.84&0.52&0.88&0.55\\
(g)&1.0&$r./o.$&0.91&0.62&0.73&0.80&0.84&0.55&0.86&0.61\\
(h)&1.0&$r./n.$ & 0.91&0.61&0.70&0.83&0.85&0.59&0.85&0.66\\
\specialrule{.1em}{.05em}{.05em}
\end{tabular}
}
\end{table}


\begin{figure}
    \centering
    \includegraphics[trim={0 0 0 1cm}, clip, width=1\linewidth]{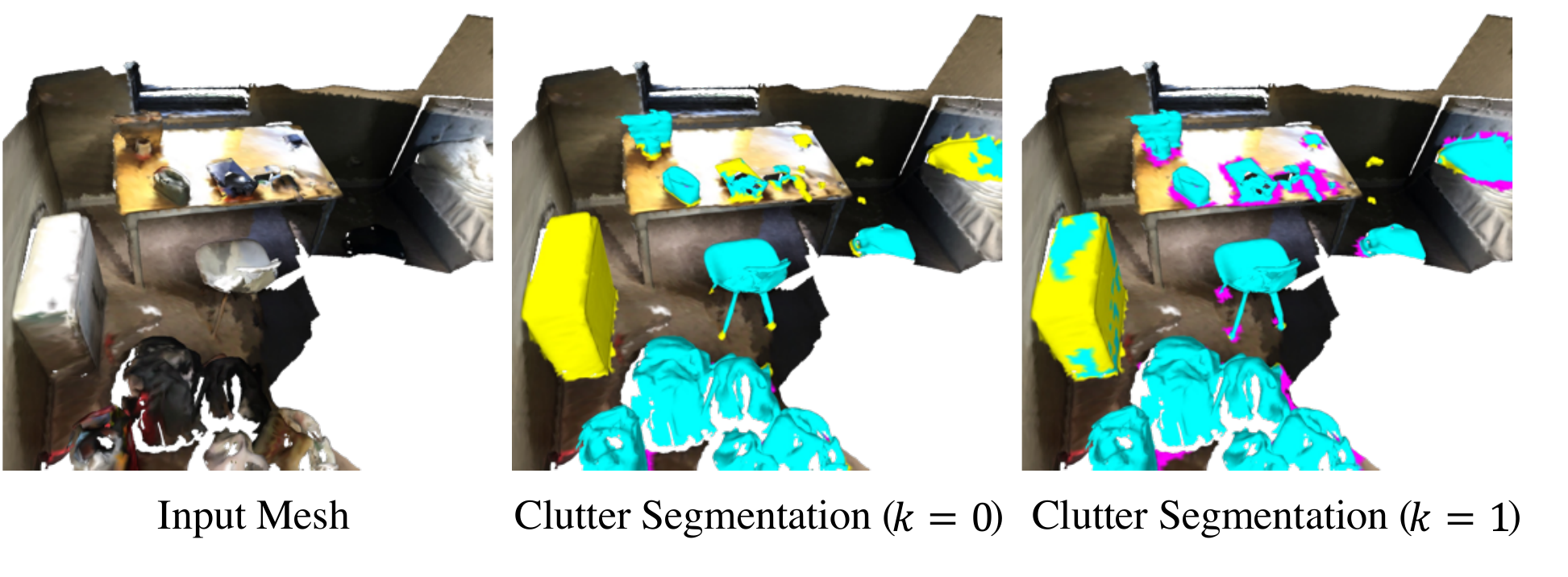}
    \caption{\textbf{Visualizations of the impact of size-balancing parameter $k$ on clutter detection.} Areas marked in cyan are true positives, magenta are false positives, and yellow are false negatives.}
    \label{fig:k}
\end{figure}
In Tab.~\ref{tab:3dseg-abl}, we investigate the effect of each strategy introduced in Sec.~\ref{subsec:clutter-seg}. For both original and manual test sets, we report IoU for non-clutter (NC) and clutter (C) classes as well as precision and recall for clutter. We will focus on the manual test set which has more accurate clutter annotations. 

We first study the effects of \textit{area-sensitive loss} by varying $k$ from 0 to 2 (a-e). All (a-e) are trained and tested on rendered views from sampled cameras. On the manual test set, we see by increasing $k$ from 0 to 1.5, IoU(C) improves by over 0.1 while IoU(NC) only fluctuates slightly. The asymmetrical change is due to the unbalanced size distribution for the two classes. Given the difficulty of segmenting small objects with very noisy labels, the substantial increase of clutter IoU from (a) to (d) proves the effectiveness of our area-sensitive loss. To better explain the increase, we further look into the precision and recall for clutter. As shown in the last two columns of Tab.~\ref{tab:3dseg-abl} (a-d), a larger $k$ (within some range 0-1.5) significantly increases the recall with a falling precision. When $k$ goes up to 2 (Tab.~\ref{tab:3dseg-abl} (e)), all metrics start to fall due to disproportionally large weights. 

The above trend in (a-d) is reflected in Fig.~\ref{fig:k}. An increased recall corresponds to the enlarged cyan (true positive) and shrunk yellow (false negative) regions. The larger pink (false positive) regions imply a drop in precision. With $k=1$, the model can segment the tiny clutter on the desk and fully segment the pillow on the top right, which is partially missed when $k=0$. The lower precision is mostly caused by false positives (pink) around clutter. This is not as harmful as expected since we originally also dilate predicted masks to remove shadows. We want to note that the currently missed predictions are very hard cases. For example, the big box on the left is very similar to a small cabinet and the raw category \textit{box} itself is a mix of clutter and non-clutter. The two small yellow missing spots on the top right are two small plug heads, and their surrounding mesh surface is very dark to recognize the clutter. 

In the original test set, we can observe both IoU(C) and IOU(NC) drop with increasing $k$. Comparing the changes of IoU(NC) in both test sets from (a) to (d), a large drop (0.94 to 0.86) is observed in the original set while the value stays the same in the manual set (0.83 to 0.83). This is because by increasing the area-sensitive strength through $k$, more clutter objects are recognized, which can be mislabeled as non-clutter in the original test set. This also verifies our previous claim (Sec.~\ref{sec:intro}) that existing datasets have very low-quality clutter labels. Despite the low-quality label, we also observe the trend of increased recall and declined precision similar to that of the manual test set.

We compare the effect of using rendered or captured images in Tab.~\ref{tab:3dseg-abl} (c, f, g, h).  All (f-h) experiments are tested on captured images. We can see that (g) trained on virtual renderings with captured camera parameters performs on par with (f) trained on original captured images. This shows that the 2D domain gap does not affect the performance. From (g) to (h), we see that the larger variations of camera pose (h) improve over using only poses on the hand-held sensor trajectories (g). Comparing (c) and (h) that use the same model trained on $r./n.$ views, we see that the model trained with virtual renderings performs better when tested on original captures (h) than on its virtual renderings (c). This is probably because testing on captured images is easier since the cameras all fall into a single trajectory.



\subsubsection{3D Inpainting}

\begin{figure}[t!]
    \centering
    \includegraphics[trim={0.5cm 0 0 0}, clip, width=1\linewidth]{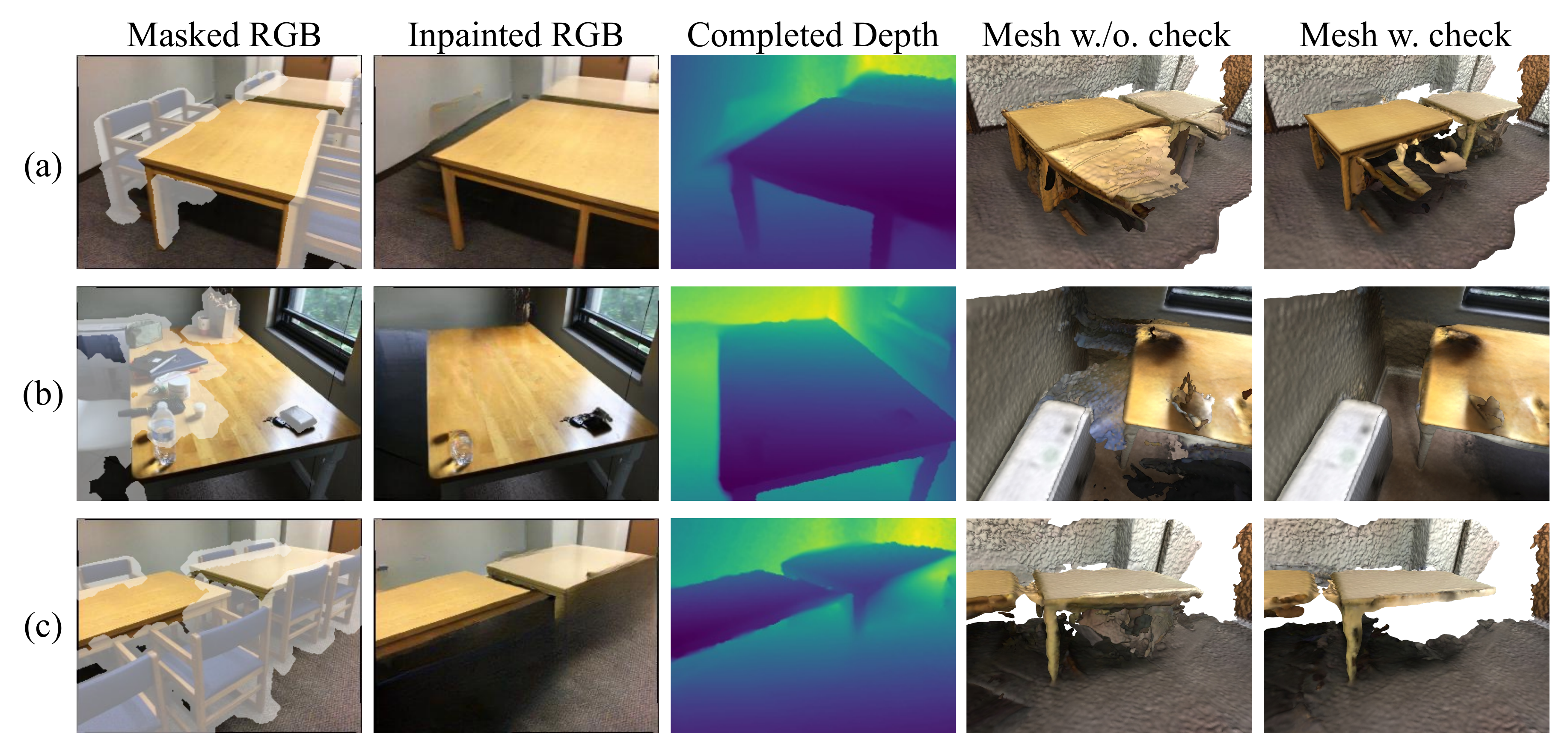}
    \caption{\textbf{Visualizations of the necessity and effect of consistency rules.} Original images (first column) are independently inpainted (second and third column), frequently leading to poor merged results (fourth column).  Our checks result in more plausible reconstructions (fifth column). The three rows illustrate the effects of (a) single-frame pruning, (b) cross-frame pruning, and (c) cross-frame voting. }
    \label{fig:ablation-inp}
\end{figure}
Despite image inpainters' great power of inpainting semantically plausible content, consistency across views is never guaranteed. In this section, we study the effect of applying different consistency rules, visualized in Fig.~\ref{fig:ablation-inp}. 

For a single frame, while the inpainted content can look plausible with the remaining pixels, the inpainted image may not conform to the original RGB image. In row (a), for instance, although the inpainted RGB completes a plausible table, these regions originally occluded by chairs should instead be the floor. If we directly reconstruct a mesh from a sequence where occluded views are inpainted differently, we will get a mesh in the fourth column. A simple yet effective solution is to discard frames where the inpainted regions have smaller depth values than the original frames (single-frame consistency in Sec.~\ref{subsec:multi-inpaint}).

Not only RGB inpainter, but inaccurate depth predictions can also cause artifacts in the reconstructed mesh. Row (b) demonstrates a case where the depth completion for the space between the desk and the wall can easily create noise. The mesh in the fourth column is what we will get without correction. We know this mesh is wrong because, from a view similar to the third row, we should be able to see the walls and floors. Such a cross-frame prior leads us to cross-frame consistency pruning: any non-inpainted regions visible in one frame should also be visible in another frame. This idea can effectively remove many artifacts and return a much cleaner mesh as shown in the fifth column.

So far, we have applied all the consistency assumptions that leverage captured images. The last step is consistency check across inpainted regions in different views. For example, the dark inpainted regions under the table can easily cause an inaccurate depth estimation. This results in the cluttered artifacts in the fourth column. By applying the cross-frame voting from Sec.~\ref{subsec:multi-inpaint}, regions with no clustered depths from different views will be dropped.


\section{Conclusion}
\label{sec:conclusion}
This paper studies the problem of detecting and removing clutter from 3D scenes, while providing plausible fillings. We started by defining clutter as frequently-moving objects and highlighting the issue of highly noisy annotations in existing 3D scene segmentation datasets. Then we presented a solution for  clutter segmentation and 3D inpainting. To better segment clutter objects with heterogeneous shapes, we developed an instance-level area-sensitive loss that significantly improves recall, particularly for small objects. Our 3D inpainting method is grounded in the key principle of decomposing the task into geometry and color (semantics). To achieve this, we project 3D segmentation onto RGB-D images to complete the color and depth in the 2D space. These 2D inpaintings are then used for mesh reconstruction after view-consistency voting and pruning. To our knowledge, this is the first work that investigates the combined problem of clutter segmentation and inpainting in 3D. While our method compares favorably with baseline methods, there remain several open problems. For instance, removing shadows before inpainting and approaches that solve this problem directly in 3D space are both intriguing avenues for further inquiry.

\paragraph{Acknowledgement} We would like to thank Intel and Google for partial funding of this work.

{\small
\bibliographystyle{ieee_fullname}
\bibliography{egbib}
}

\clearpage
\appendix

In this supplementary material, we first explain in detail the datasets used to evaluate 3D segmentation and 3D inpainting in in Sec.~\ref{sup_sec:data}. Then we describe more implementation details in Sec.~\ref{sup_sec:implement}. Finally, we present additional results of our method in Sec.~\ref{sup_sec:result}.
\section{Datasets}
\label{sup_sec:data}
\subsection{Clutter Segmentation}
\begin{figure*}
    \centering
    \includegraphics[trim={0 1.2cm 0 1.2cm}, clip,height=0.93\textheight]{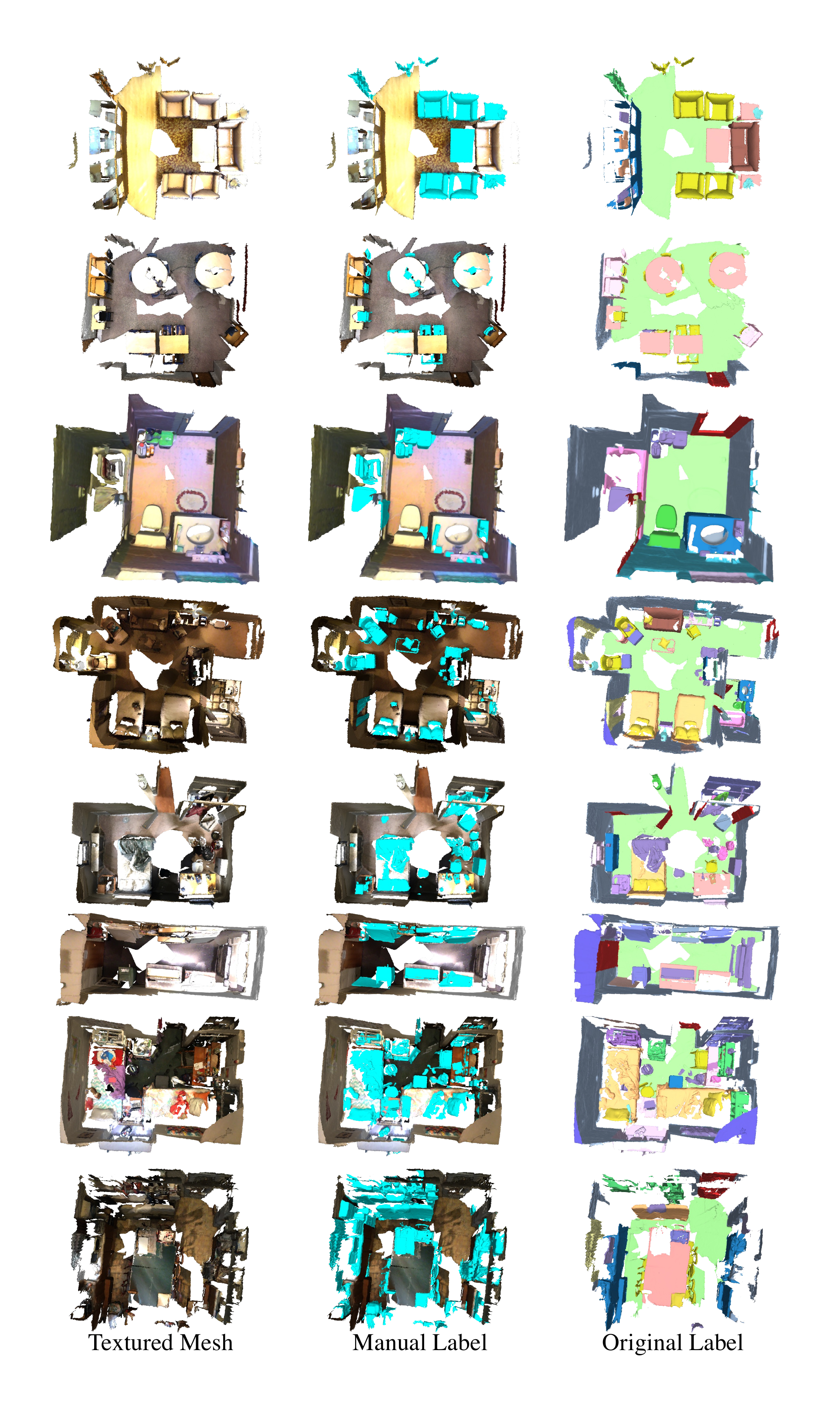}
    \caption{\textbf{Clutter segmentation manually-labeled evaluation dataset.} We show the bird eye views of all eight scenes (scaled, rotated, and ceiling surfaces removed for better view) that we manually labeled to evaluate the clutter segmentation model of our method and the BPNet~\cite{hu-2021-bidirectional} baseline.}
    \label{fig:manual-data}
\end{figure*}
We select eight representative scenes from ScanNet test set to cover as many as the scene types in the dataset as possible. We use the annotation tool from ScanNet~\cite{dai2017scannet} to annotate vertices on clutter objects. In Fig.~\ref{fig:manual-data}, we show the bird eye views of all eight scenes that we manually labeled to evaluate the clutter segmentation models. In particular, the manual test set covers a wide range of cleanness levels from rooms with very few common clutter objects to highly cluttered rooms. The covered room types include study room, meeting room, bedroom, bathroom, living room, studio, and craft room. As we can see from the third column, in original ScanNet labels, some clutter objects are unlabeled (\eg, second row, objects on the round table) and some objects are mislabeled (\eg, last row, clutter objects in the dark blue closet). In comparison, our manually annotated dataset correctly labels all clutter in the scenes.
The time required for each scene varies from 30 minutes (only for extremely clean scenes) to hours. This also demonstrates that it is impractical to annotate a dataset of thousands of scenes with precise clutter/non-clutter labels.

For all experiments, we sample once every 5 frames from the raw captured sequences to reduce redundancy and improve efficiency. To render virtual views for clutter segmentation, we use cameras with fixed intrinsics. We randomly put cameras at a distance of 2-5 meters from the mesh surface looking at the scene to render views for training. 




\subsection{3D Inpainting}\label{sec:dataset-inp3d}
The metrics used to quantitatively evaluate 3D inpainting (main Tab. 2) 
requires ground truth, which is not available in any existing datasets. Therefore, we construct a synthetic evaluation set. 
We use the official test split from Matterport3D~\cite{chang2017matterport3d} which originally has 391 rooms. For each room, we identify regions that belong to non-clutter categories and have a normalized normal $n=(n_x, n_y, n_z)$ where $|n_z|=max(|n_x|, |n_y|, |n_z|)$ ($z$ is pointing up).
In the meantime, we cut instances that belong to the class pure clutter from the scenes and drop those instances on the identified non-clutter regions. 
We find the convex hull of the dropped clutter and any surface of the room that is contained within the convex hull is removed to create holes. In other words, the original room is the ground truth room after 3D inpainting, and the room with introduced holes is the input to different 3D inpainting methods. 

Both baseline methods use TSDF representation as input and output. To create the input, we mask out the TSDF and color values for the voxels that are within the convex hull. This is very similar to how FF~\cite{jheng2022free} creates free-form holes in their methods.

For each hole of the room, we find a view from the dataset where the hole is placed as close to image center as possible. We also find other views that have at least 1000 pixel overlap with the centered view.  We project the clutter onto all the above views to create holes on RGB-D data. We then use them to run our RGBD inpainting and mesh reconstruction. 

Below we explain how we compute 3D and 2D metrics once the inpainted 3D is obtained (from either our method or baselines). To evaluate the proposed method's 3D performance, we compare Chamfer Distance between our inpainted mesh and groundtruth mesh reconstructed (using PSR) using original (unmasked) views. To evaluate baselines' 3D performance, we compare Chamfer Distance between marching cubes mesh from predicted TSDF and marching cubes mesh from groundtruth TSDF. This is fairer than using mesh reconstructed from PSR as groundtruth for baseline methods (using PSR results as groundtruth would result in worse results than using marching cubes results from TSDF, so we don't use PSR as groundtruth for baselines). To evaluate 2D performance, we render from the centered view and compute the metrics between the ground truth renderings (without dropped clutter) and renderings from our method or baselines.

\section{Implementation Details}
\label{sup_sec:implement}
\begin{table}[t]
\caption{\textbf{Runtime analysis.} We compute average Timing based on scenes that have approximately 200 frames and a $150,000$-vertex input mesh from original scan data. All GPUs are NVIDIA TITAN X with 12 GB memory. We can see that it takes slightly over 30 min to process a 200-frame RGB-D sequence, and the majority of the time is spent on consistency checks and surface reconstruction.  }
\label{tab:timing}
\centering
\resizebox{1\linewidth}{!}{
\begin{tabular}{lcccccc}
\specialrule{.2em}{.1em}{.1em}
Step&Time & Device&Output size\\
\hline
Clutter Segmentation&2s&GPU& $150,000$ points\\
Mask Projection&30s&GPU& 200 images\\
Image Inpainting&50s&GPU& 200 images $320\times240$\\
Depth Completion&80s &GPU& 200 images $304\times228$\\
Consistency Checks&18min &GPU& 200 images $304\times228$\\
Surface Reconstruction&11min&CPU &1,500,000 vertices\\
\specialrule{.1em}{.05em}{.05em}
\end{tabular}
}
\end{table}
We show in Tab.~\ref{tab:timing} the average timing of each step for a 200-frame sequence. Specifically, we compute average Timing based on scenes that have approximately 200 frames and a $150,000$-vertex input mesh from original scan data. All GPUs are NVIDIA TITAN X with 12 GB memory. We can see that it takes slightly over 30 min to process a 200-frame RGB-D sequence, and the majority of the time is spent on consistency checks and surface reconstruction. 

We find empirically Cross Entropy loss for two classes works better than Binary Cross Entropy loss.

\subsection{Clutter Definition}
Our focus is to propose a new type of scene segmentation based on shared properties (\eg, across clutter) rather than common benchmark semantic categories. Therefore, a concrete clutter definition is crucial to test our method but the exact form can be very flexible.

The definition we chose for our result is as follows: an uninstalled object is considered clutter if it's moved at any time within 3 months and with probability $>95\%$ so that the IoU between the bounding boxes before and after the movement is $<0.5$.  We want to note that although we need a concrete definition to run the experiments, our proposed techniques for clutter segmentation and 3D inpainting are general and not specifically tailored to work for only one definition. For example, one can easily extend the current clutter definition to be with variable time length (\eg, 1 or 9 months) without affecting the effectiveness of our proposed methods.

\subsection{Clutter Segmentation}
The model is implemented in PyTorch, trained with batch size 16 on two NVIDIA TITAN X with 12 GB memory. Following BPNet~\cite{hu-2021-bidirectional}, we use SGD optimizer with base learning rate of 0.01 and employ a poly learning rate scheduler with the power set to 0.9. Momentum and weight decay are set to 0.9 and 0.0001, respectively. The model is trained in total for 100 epochs.

 During testing, we run the inference on the same mesh multiple times (each time with different camera views) until all camera views have been used. 
 We accumulate the model's predicted class probability for the final 3D segmentation results.

After projecting the 3D segmentation masks onto RGBD images, we further dilate the 2D masks with 6 iterations (pixels) to account for the misalignment between 3D reconstruction and 2D captures. The final masks are applied to color and depth images to remove regions with clutter.

\subsection{Depth Completion}

As introduced in main paper, we adopt NLSPN~\cite{park2020non} that performs non-local depth propagation for image-guided depth completion. The model was originally designed for the task of sparse depth completion that predicts the complete depth map from hundreds of sampled pixels. One of the method's key components is the deformable convolution~\cite{zhu2019deformable} whose effective sampling field can be the entire image. Therefore, despite its different original goals, this method is still proper for our usage of completing the dense map with (potentially big) holes. We retrain the model on our depth maps with synthetic holes for depth completion. We explain below in detail how we prepare the training data.

We use captured depth data with modifications. Since there is no ground truth depth map before and after clutter removal, we need synthesize training data by ourselves. To do so, we start from a captured depth map $d_1$. We assume $m_1$ is the mask for the coarse clutter/non-clutter groupings (main paper Sec. 3.2.1) of $d_1$. We use the clutter mask $m_2$ from a second view to mask out regions on $d_1$. In the meantime, we make sure that the regions originally masked as clutter in $m_1$ are not masked out. By copying and pasting $m_2$ from another view, we make sure that the masks during training have realistic boundaries of clutter objects. By not masking clutter from $m_1$, we guarantee that the depth completion model does not need to hallucinate clutter objects out of the hole areas. And this is exactly the expected behaviors of a depth completion model. We note that since the clutter/non-clutter grouping is very noisy, the training data created as described above still contains noise. However, we empirically find the model trained this way already works much better than using pre-trained model weights trained on the sparse-to-dense depth completion task.

The model is implemented in PyTorch, trained with batch size 12 on one NVIDIA TITAN X with 12 GB memory. Following NLSPN~\cite{park2020non}, we use Adam optimizer with $\beta_1=0.9,\beta_2=0.999$ and initial learning rate of 0.001. The model is trained in total for 20 epochs. We set the propagation time of each forward pass as 18.
\section{More Results}
\label{sup_sec:result}

\subsection{Ablation on Segmentation Loss}

\begin{table}[t]
\caption{\textbf{Ablation on segmentation loss.} We compare our area-sensitive loss with original Cross Entropy (CE), balanced Cross Entropy (BCE)~\cite{cui2019class}, median frequency loss (MF)~\cite{eigen2015predicting}, and Focal Loss (FL)~\cite{lin2017focal}.  Note: hyper-parameters were set for other methods ($\beta$ in BCE and $\gamma$ in FL) using best practices suggested in the original papers.}
\label{tab:loss-ablation}
\addtolength{\tabcolsep}{-2pt}
\centering
\resizebox{0.5\linewidth}{!}{
\begin{tabular}{lcccccc}
\specialrule{.2em}{.1em}{.1em}
Method  &\multicolumn{3}{c}{Manual Test Set (clean)} \\
       & IoU(NC) & IoU(C) & mIoU \\
\hline
CE&0.83&0.48&0.66\\
BCE&0.74&0.38&0.56\\
FL&0.83&0.45&0.64 \\
MF&\textbf{0.84}&0.54&0.69  \\
Ours&\textbf{0.84}&\textbf{0.58}&\textbf{0.71}\\
\specialrule{.1em}{.05em}{.05em}
\end{tabular}
}
\end{table}

Tab.~\ref{tab:loss-ablation} shows an ablation study comparing different losses designed to combat imbalance.  Our per-instance area-sensitive loss performs the best.  BCE~\cite{cui2019class} and FL~\cite{lin2017focal} perform worse than the CE baseline for clutter segmentation.  MF~\cite{eigen2015predicting} improves over CE by re-weighting classes based on cumulative class surface area, but it is worse than ours.  Since clutter usually comprises many small object instances, per-instance weighting is advantageous for this task.

\subsection{Ablation on Consistency Checks} 

\begin{table}[t]
\caption{\textbf{Ablation on consistency checks.} sp, cp, cv stand for single-frame pruning, cross-frame pruning, and cross-frame voting, respectively.}
\label{tab:consistency-ablation}
\addtolength{\tabcolsep}{-3pt}
\centering
\resizebox{1\linewidth}{!}{
\begin{tabular}{lccccccccc}
\specialrule{.2em}{.1em}{.1em}
ID &sp&cp & cv &L1($\downarrow$)& L2($\downarrow$)& PSNR($\uparrow$)& SSIM($\uparrow$)&LPIPS($\downarrow$)&CD(cm)($\downarrow$)\\
\hline
a&&&&0.034&0.075&21.820&0.860&0.201&9.013\\
b&\checkmark&&&0.029&0.051&22.403&0.873&0.189&5.245  \\
c&\checkmark&\checkmark&&0.023&0.043&22.965&0.880&0.171&2.612\\
d&\checkmark&\checkmark&\checkmark&0.019&0.035&23.820&0.891&0.150&0.931\\
\specialrule{.1em}{.05em}{.05em}
\end{tabular}
}
\end{table}

Tab.~\ref{tab:consistency-ablation} studies the contribution of each consistency check on reconstruction for 100 Matterport3D test scenes (Sec.~\ref{sec:dataset-inp3d}). Every consistency check improves the 3D inpainting, especially the mesh reconstruction quality.

\subsection{Ablation on $\alpha,\beta$}

\begin{figure}[t]
\begin{center}
   \includegraphics[width=1\linewidth]{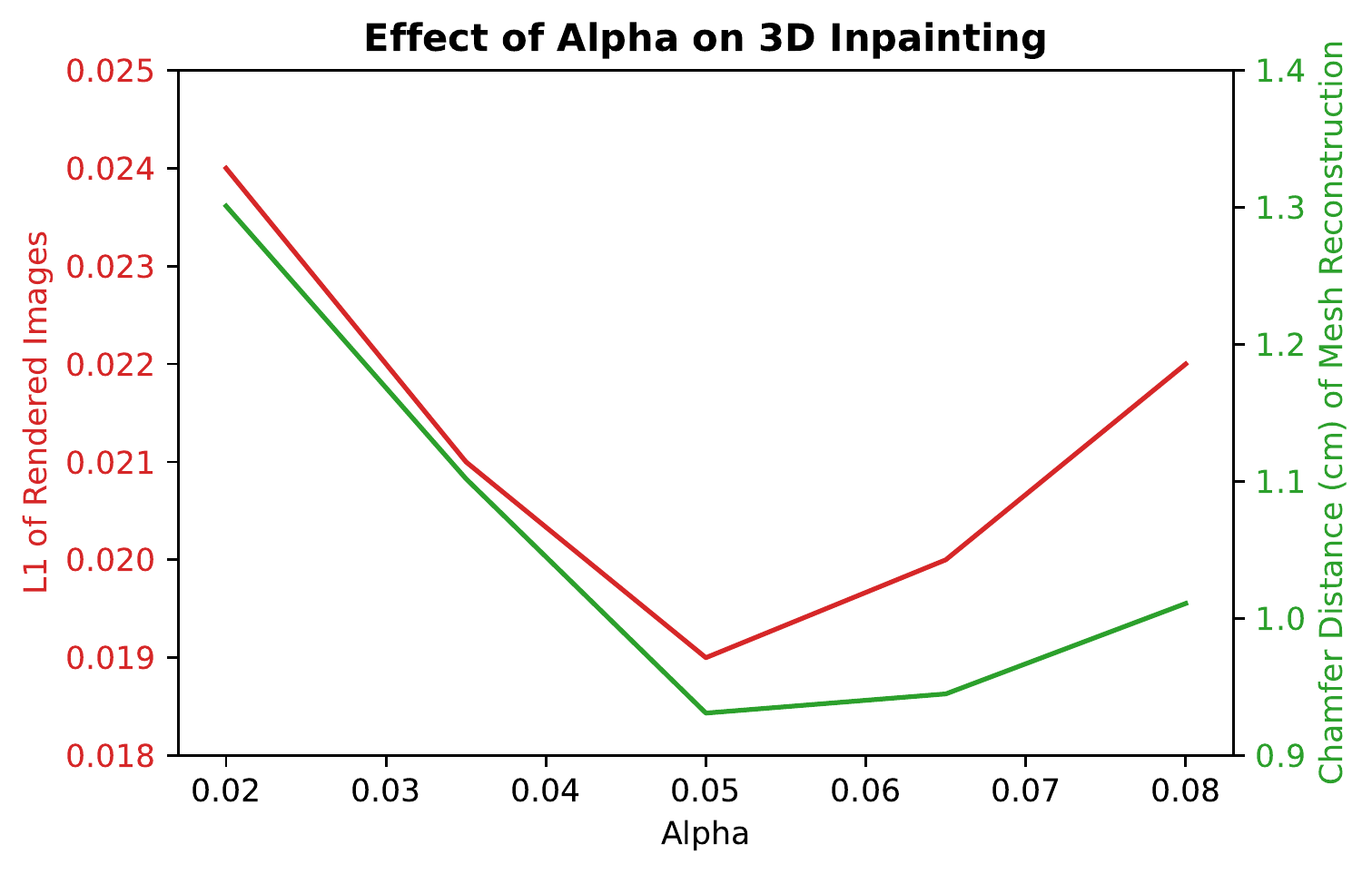}
   \includegraphics[width=1\linewidth]{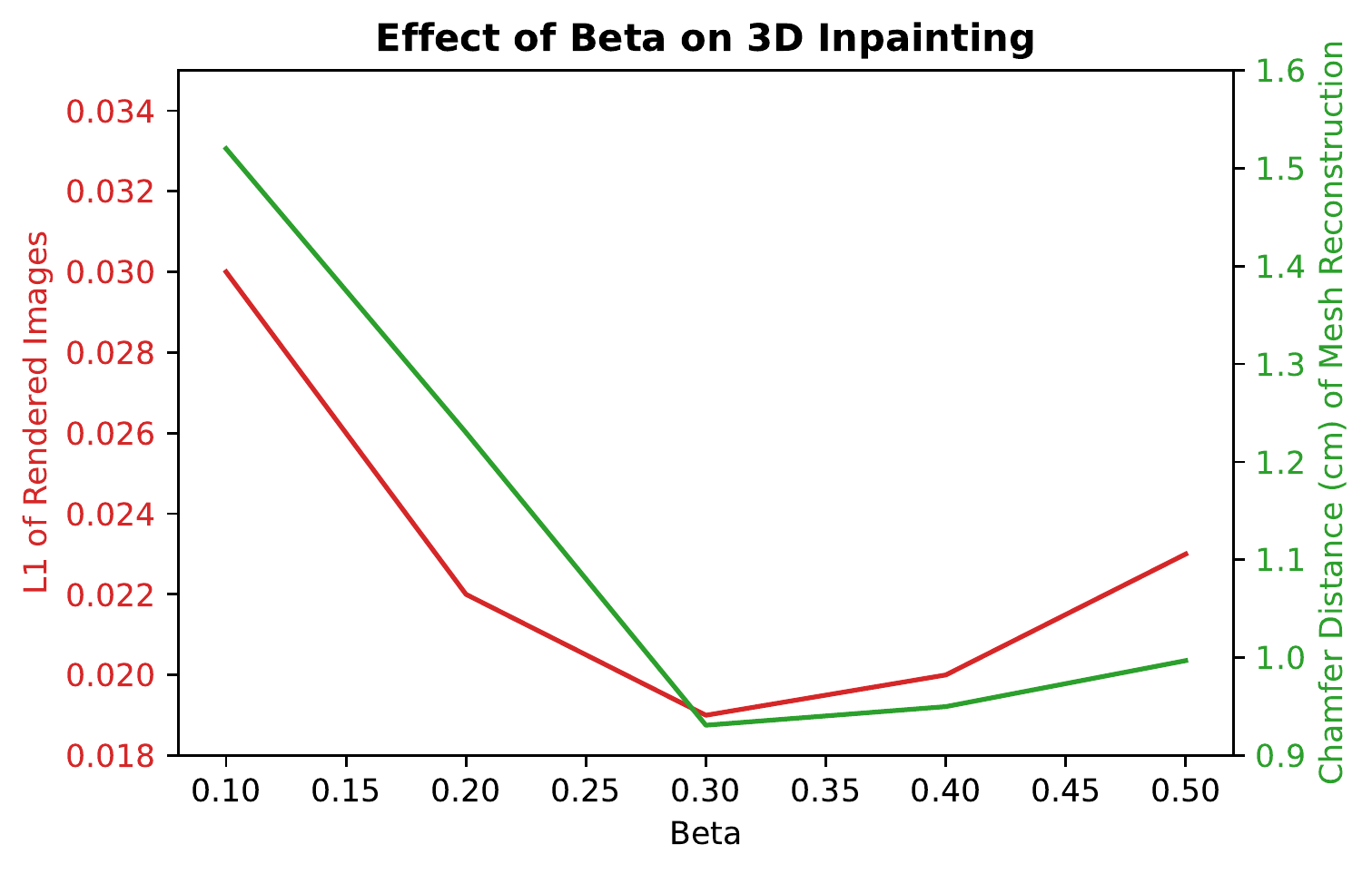}
\end{center}
   \caption{\textbf{Ablation on $\alpha$, $\beta$ on 3D geometry and rendered image}.}
\label{fig:ab-ablation}
\end{figure}

Fig.~\ref{fig:ab-ablation} studies the effects of varying $\alpha$ and $\beta$ on 3D inpainting.
$\beta$ is relatively high because there are limited overlapping views for the same region; a lower $\beta$ increases the sensitivity to noise.

\subsection{Qualitative Results}
\begin{figure*}[t!]
    \centering
    \includegraphics[trim={0 0 0 0.5cm}, clip,width=0.95\linewidth]{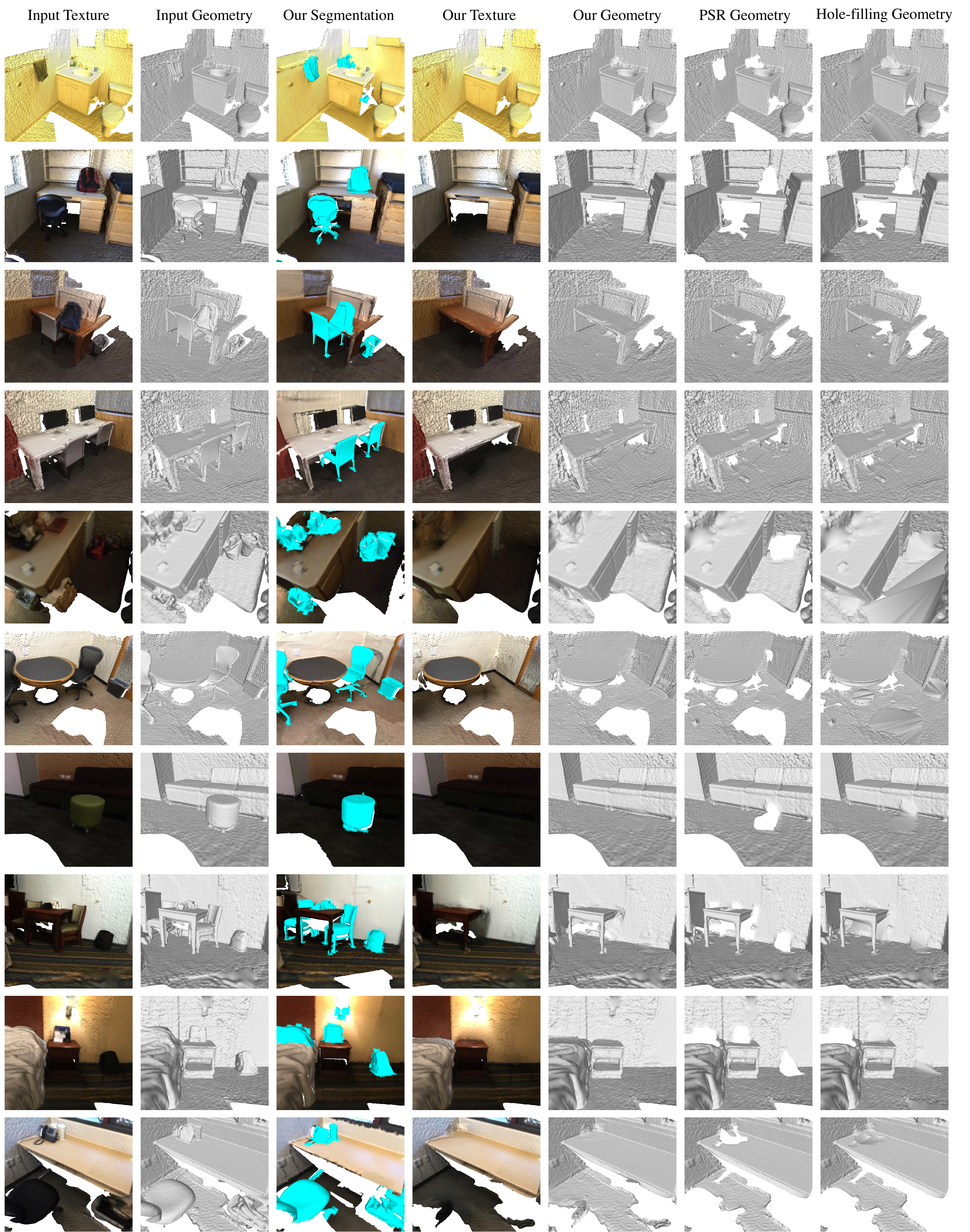}
    \caption{\textbf{Additional results for automatic clutter removal and 3D inpainting.}  Each row shows a rendered view of an input 3D scene, together with our clutter segmentation and a rendered view of the scene with inpainted color and geometry.  We compare the latter to removing clutter and filling the resulting holes with either Poisson Surface Reconstruction (PSR) or a triangulation-based hole-filling algorithm.}
    \label{fig:add-result}
\end{figure*}
\begin{figure*}[t]
    \centering
    \includegraphics[trim={0 0 0 0.5cm}, clip,width=0.95\linewidth]{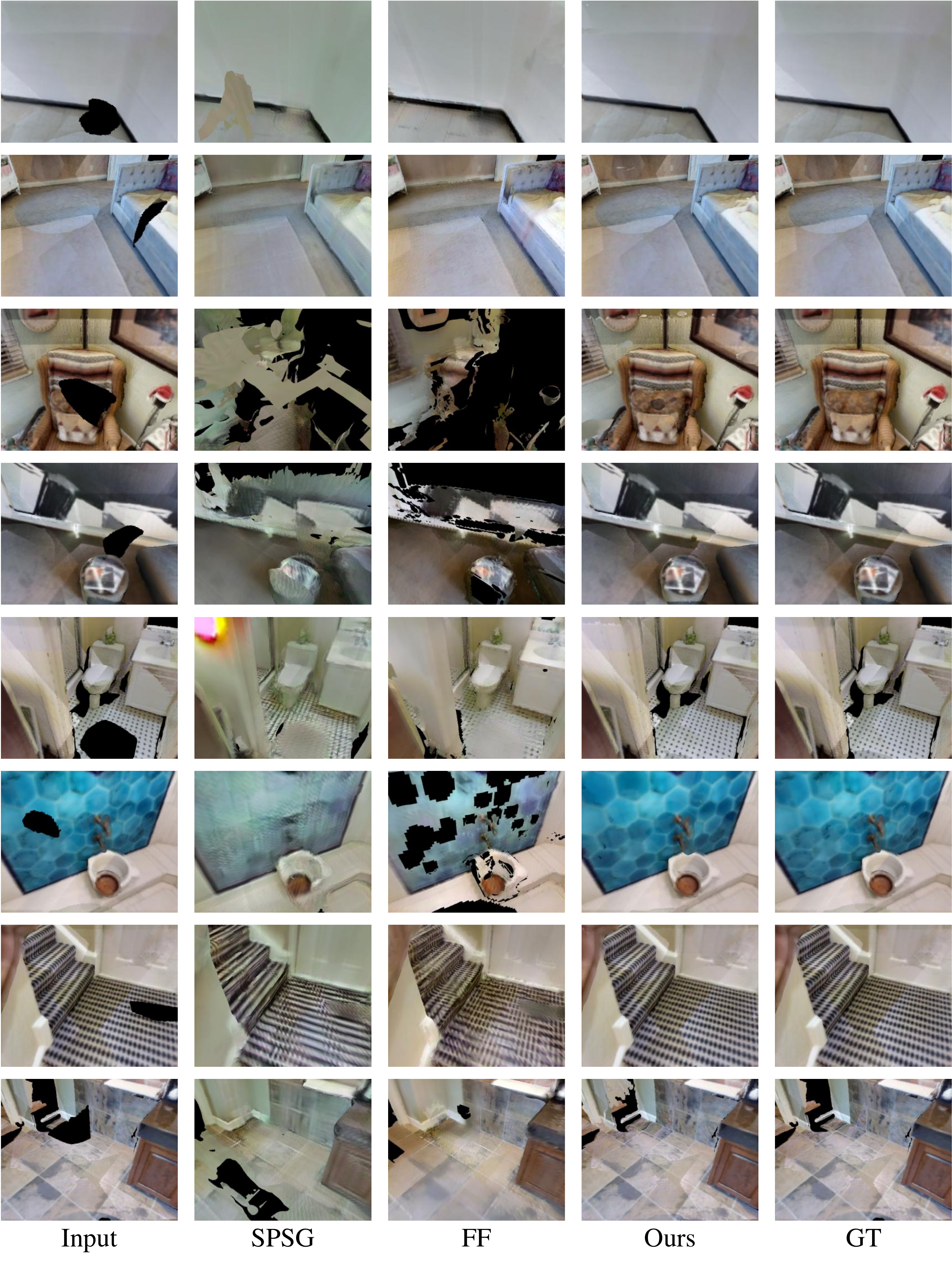}
    \caption{\textbf{Additional results for 3D inpainting.} The first column shows a rendered view of an input 3D scene, The second and third rows are results from baseline SPSG and FF. The fourth column is results of proposed method. The last column is ground truth mesh.}
    \label{fig:add-result-inp}
\end{figure*}
\begin{figure*}[t]
    \centering
    \includegraphics[trim={0 0 0 0.5cm}, clip,width=0.95\linewidth]{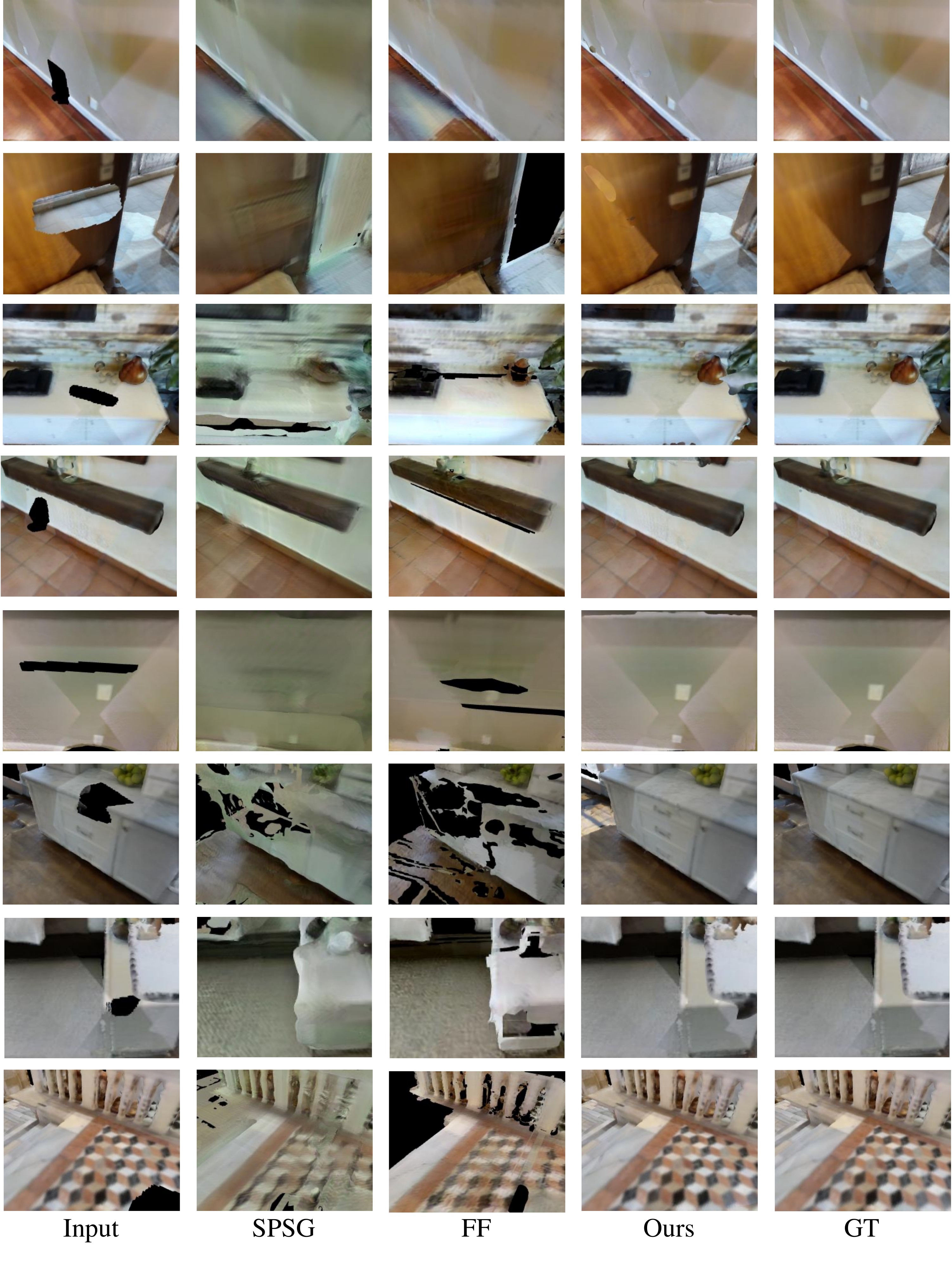}
    \caption{\textbf{Additional results for 3D inpainting.} The first column shows a rendered view of an input 3D scene, The second and third rows are results from baseline SPSG and FF. The fourth column is results of proposed method. The last column is ground truth mesh.}
    \label{fig:add-result-inp2}
\end{figure*}
In Fig.~\ref{fig:add-result}, we show additional qualitative results for our entire pipeline of object removal and 3D inpainting, where we compare with baselines of 3D inpainting. The layout is the same as in the main paper. Alongside input mesh texture and geometry, we first show our predicted 3D clutter segmentation. Then we show the texture and geometry of our final inpainted mesh. In the last two columns, we compare with the geometry from Poisson surface reconstruction (PSR)~\cite{kazhdan2006poisson} and a hole-filling algorithm. For fair comparison, the input meshes (first two columns) are also reconstructed using the same Poisson surface reconstruction settings as described in main paper Sec. 3.3.3. The mesh used to visualize segmentation (third row) is from original ScanNet dataset. This is a much smoothed and simplified version than the mesh reconstructed using Sec. 3.3.3.

In Fig.~\ref{fig:add-result-inp} and Fig.~\ref{fig:add-result-inp2}, we show additional results of 3D inpainting compared with two baselines, SPSG~\cite{dai2021spsg} and FF~\cite{jheng2022free}. We can see that our method outputs results of higher resolution than prior work that uses volumetric representations. The proposed method also fills in the hole with more coherent texture and geometry compared to baselines.

\end{document}